\documentclass[]{bytedance}
\usepackage{float}

\DeclareFontShape{T1}{bytesans}{b}{n}{<-> s * [1] seed/bytesans}{}
\DeclareFontShape{T1}{bytesans}{bx}{n}{<-> s * [1] seed/bytesans}{}
\title{FlowAct-R2: Beyond Talking Avatar via Streaming Multimodal References and Proactive Agent Planning}

\author{
    \centerline{Ziyao Huang$^*$ \quad Zhengkun Rong$^*$ \quad Shiyang Qin$^*$ \quad Shuang Liang$^*$}
    \centerline{Wentao Hu$^*$ \quad Yuxuan Luo$^{*\dagger}$ \quad Yuan Zhang \quad Mingyuan Gao}
}
\affiliation[]{Bytedance Intelligent Creation}
\contribution[]{$^*$Core Contributors \qquad $^\dagger$Corresponding Author}

\abstract{
We present FlowAct-R2, a framework for interactive humanoid video generation that combines continuous multimodal control with proactive agent planning. Our method consists of two coupled components. First, a Streaming Multimodal Reference Diffusion Transformer adapts the pretrained Seedance 2.0 Mini reference-to-video backbone to accept rolling action prompts, streaming audio, and dynamically updated image, audio, and video references. Video-driven rotary positional embeddings align reference chunks with the generation timeline, while reference-plus-image conditioning and partially noised historical motion frames preserve appearance and avoid accumulated drift. Second, a Proactive Interaction Agent separates pre-online planning from online scheduling and response: it prepares a persona, a long-horizon agenda, and reusable multimodal skills in advance, then autonomously schedules behaviors, responds to audience input, and handles interruptions during a live session. FlowAct-R2 supports real-time 720p generation and hour-scale streaming across entertainment streaming, live shopping, video chatting, and live vlogging.
}
\date{September 29, 2026}
\checkdata[Project Page]{\url{https://bone-11.github.io/Flowact-R2/}}
\checkdata[Hugging Face Space]{\url{https://huggingface.co/spaces/ProAudience/FlowAct-R2}}

\hypersetup{
    pdftitle={FlowAct-R2: Beyond Talking Avatar via Streaming Multimodal References and Proactive Agent Planning},
    pdfauthor={Ziyao Huang, Zhengkun Rong, Shiyang Qin, Shuang Liang, Wentao Hu, Yuxuan Luo, Yuan Zhang, Mingyuan Gao},
    pdfsubject={Streaming humanoid video generation and proactive interaction}
}

\begin{document}
\maketitle
\section{Introduction}
\label{sec:intro}

Digital humans are emerging as practical tools for interactive applications, yet most existing systems remain confined to a single scenario, such as dialogue~\citep{li2024ditto,zhu2024infp,huang2025liveavatar,xie2025xstreamer}, talent performance~\citep{zhang2026liveanimate,zhang2026vidus2}, or live commerce and product demonstration~\citep{rao2026streamhoi,vivix2026a1}. To support open-ended live interaction, however, a digital human should go beyond a collection of task-specific generators and behave more like a real human streamer: it should maintain a coherent persona and long-term context, proactively organize what to do next~\citep{wooldridge1995agents,park2023generativeagents,yao2022react,wang2023voyager}, respond to unexpected user input without abandoning the ongoing activity~\citep{thinkingmachines2026interaction,huang2026duplexomni}, and coordinate speech, body motion, object interaction, and scene-level actions.

This requires both a generator capable of continuously realizing changing multimodal behaviors and an agent responsible for planning, scheduling, interrupting, and resuming them throughout a long-running session. Recent systems have advanced continuous humanoid video generation, real-time audiovisual interaction, and interactive world simulation~\citep{wang2026flowactr1,huang2026wanstreamer,pixverse2026r2}, but two coupled challenges remain. The generator needs to incorporate dynamically changing image, audio, and video references while preserving identity and temporal continuity. Meanwhile, the agent needs to prepare complex activities in advance and make latency-sensitive decisions online, allowing unexpected user inputs to redirect ongoing behavior without disrupting the session.

\begin{figure}[t]
    \centering
    \includegraphics[width=\linewidth]{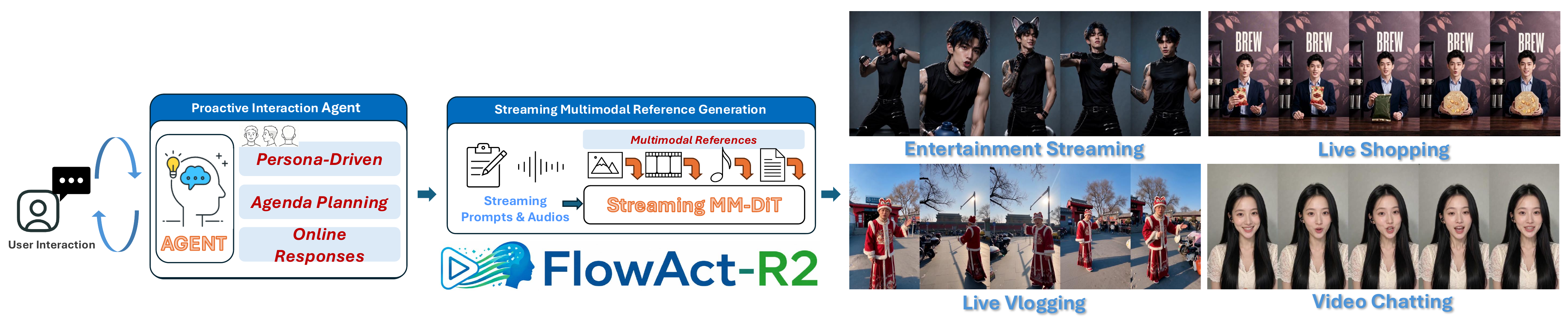}
    \caption{FlowAct-R2 couples proactive interaction with streaming multimodal reference generation. Our demonstrations cover entertainment streaming, live shopping, live vlogging, and video chatting.}
    \label{fig:teaser}
\end{figure}

To address these coupled challenges, we present FlowAct-R2, which pairs a streaming multimodal reference generator with a proactive interaction agent (Figure~\ref{fig:teaser}). This design provides three main capabilities:

\begin{itemize}
    \setlength{\itemsep}{2pt}
    \setlength{\parskip}{0pt}
    \item \textbf{Streaming Multimodal Reference Generation.} By extending diffusion forcing from chunkwise generation to streaming multimodal references, FlowAct-R2 enables real-time video generation with rolling prompts and multimodal references, including continuously updated images, audio, and video, and supports hour‑level ultra‑long video generation with 720p.
    \item \textbf{Proactive Interaction Agent.} To enable digital humans to autonomously drive their behavior when no external input is present, while seamlessly responding when interactions arise, FlowAct-R2 introduce an agent workflow with pre-online planning and online scheduling and response, which supports persona-driven behaviors, interruption handling, and skill execution for autonomous live interactions.
    \item \textbf{Rich Behaviors Across Diverse Live Scenarios.} FlowAct-R2 supports real-time text and audio interactions across four representative scenarios—video chatting, live shopping, entertainment streaming, and interactive gaming—generating not only conversational responses but also expressive behaviors and task-oriented actions beyond conventional question answering.
\end{itemize}

\section{Method Overview}
\label{sec:method}

\begin{figure}[htbp]
    \centering
    \includegraphics[width=\linewidth]{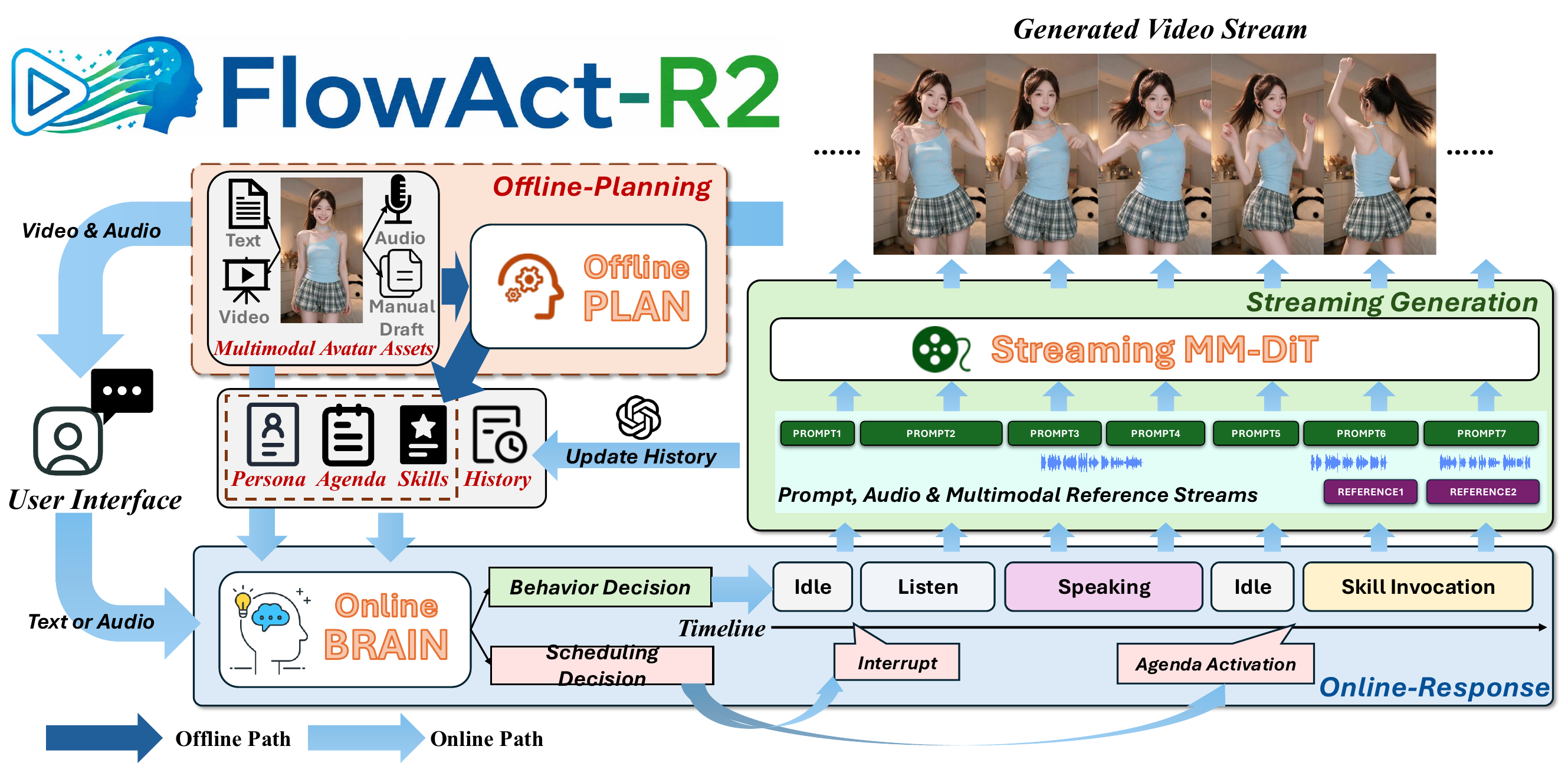}
    \caption{System overview. Offline planning organizes multimodal avatar assets into a persona, an agenda, and reusable skills. The online agent makes behavior and scheduling decisions using audience inputs and history. The resulting prompt, audio, and reference streams condition continuous video generation.}
    \label{fig:method}
\end{figure}

FlowAct-R2 has two coupled components: a Streaming Multimodal Reference Diffusion Transformer (DiT) that produces video, and a Proactive Interaction Agent that determines the behavior and its schedule. DiTs provide the underlying Transformer-based diffusion architecture~\citep{peebles2022dit}. As shown in Figure~\ref{fig:method}, offline preparation provides the persona, agenda, and skill assets. Online scheduling combines these resources with audience events and interaction history, then supplies changing conditions to the generator.

\subsection{Streaming Multimodal Reference Diffusion Transformer.}
The generator adapts the pretrained Seedance 2.0 Mini reference-to-video (R2V) backbone, retaining its reference preservation capability. Diffusion Forcing assigns independent noise levels to sequence tokens~\citep{chen2024diffusionforcing}. FlowAct-R2 uses this formulation for autoregressive streaming inference while preserving the backbone's bidirectional spatiotemporal modeling. Conditioning also becomes a stream: action prompts, driving audio, and image, audio, or video references can be updated or switched as the session proceeds. This supports changes such as presenting a new product or invoking a reference-guided performance within an ongoing stream.

Rotary position embedding (RoPE) encodes positional relationships through rotations~\citep{su2021roformer}. FlowAct-R2 uses a video-driven variant to align incoming reference chunks with the corresponding generation timeline. This provides temporal alignment when references arrive or change during generation. Explicitly trained reference-plus-image conditioning, denoted $(R+I)2V$, anchors identity and appearance. The model additionally conditions on partially noised historical motion frames to improve robustness to imperfect generated histories and reduce error accumulation. Specialized audio-driven training supports synchronization of speech, lip movements, expressions, and head poses, while coarse-to-fine inference progressively restores visual details.

\subsection{Proactive Interaction Agent.}
Proactive agent planning follows a two-stage design that moves long-horizon preparation outside the latency-sensitive interaction loop while retaining online responsiveness.

\subsubsection{Offline planning.}
The agent derives a persona from multimodal character assets, including text, images, audio, video, and manually prepared material. It organizes the session into a long-horizon agenda and compiles complex behaviors into reusable multimodal skills. The persona specifies how the character should behave, the agenda supplies activities to pursue, and the skills provide prepared resources for execution. This moves session organization and complex behavior preparation out of the immediate interaction loop.

\subsubsection{Online scheduling and response.}
During a session, the agent combines the prepared agenda and skills with audience text or audio events and summarized interaction history. It makes a behavior decision about what to perform and a scheduling decision about when and how to perform it. Depending on the context, a behavior can be initiated, overlaid, deferred, interrupted, or resumed. This permits the avatar to continue agenda-driven activity without continuous prompting and to respond when an audience event arrives.

The schedule is expressed through idle, listening, speaking, and skill-execution periods. These periods supply rolling prompts, audio, and multimodal reference updates to the DiT. Interaction history is updated as the session evolves, giving subsequent decisions access to prior context. The agent and generator therefore operate at complementary levels: the agent manages session behavior and timing, while the generator realizes those decisions as a continuous video stream.

\section{Results}
\label{sec:results}

\subsection{Streaming and application demonstrations.}
\begin{figure}[H]
    \centering
    \includegraphics[width=\linewidth,height=0.59\textheight,keepaspectratio]{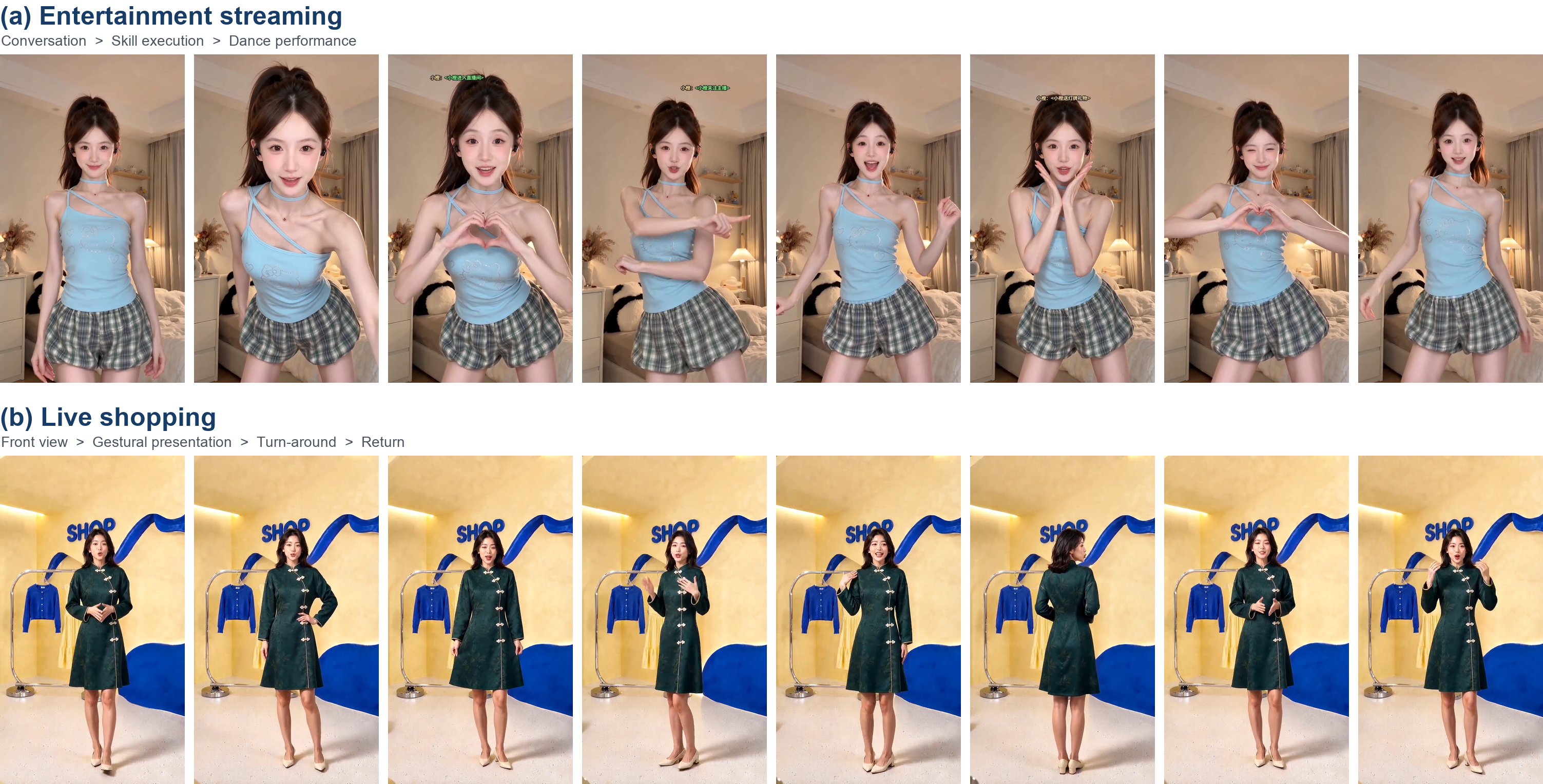}
    \caption{Entertainment and live shopping results. (a) Selected poses and gestures during a dance performance. (b) A continuous presentation of one garment, including front-facing explanation, expressive gestures, and a rear-view demonstration. Each row contains eight full frames from one video in temporal order.}
    \label{fig:performance-shopping}
\end{figure}

\begin{figure}[H]
    \centering
    \includegraphics[width=\linewidth,height=0.59\textheight,keepaspectratio]{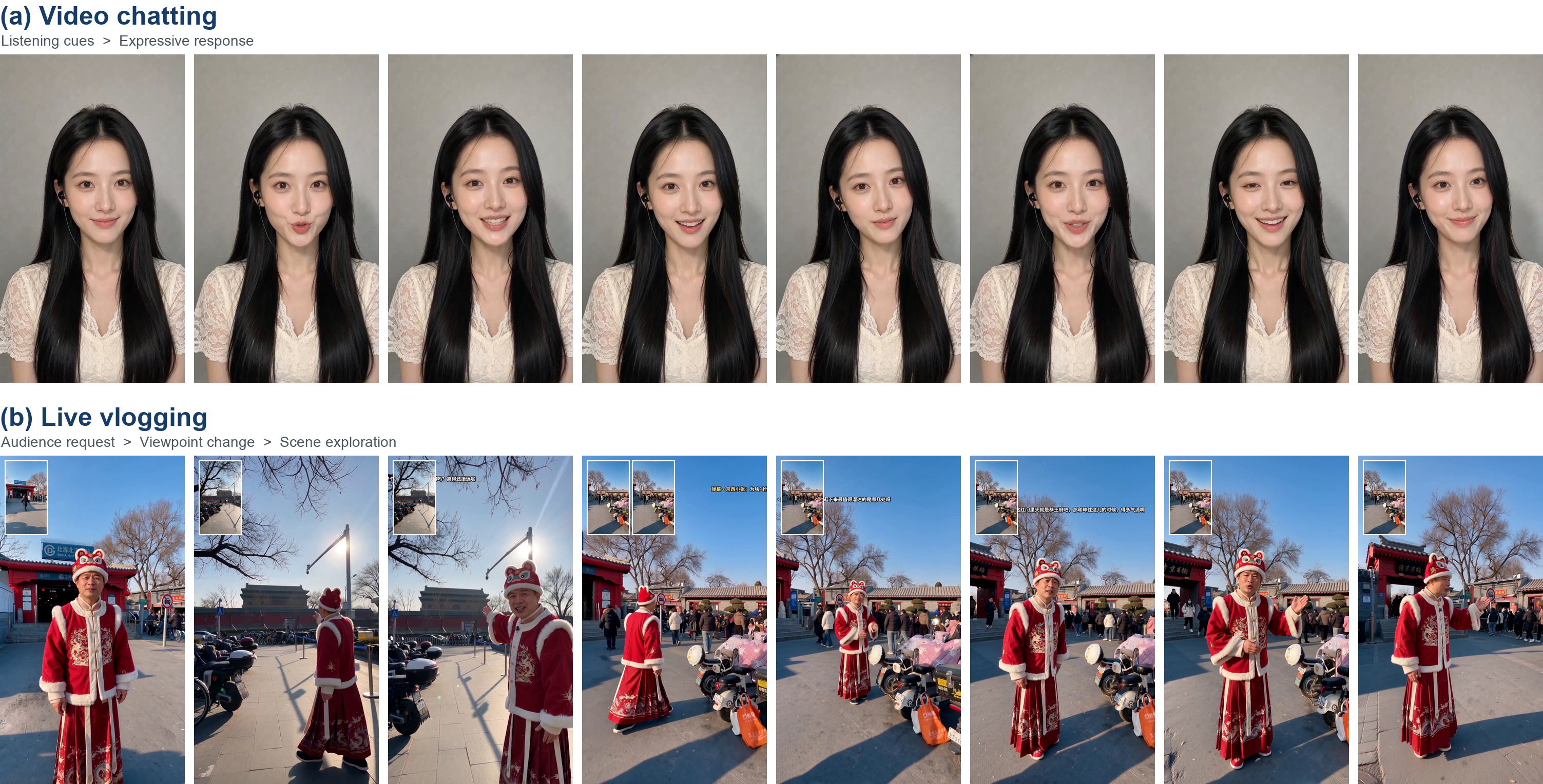}
    \caption{Video chatting and live vlogging results. (a) Facial expressions and gestures during a conversation. (b) Viewpoint and scene changes during an outdoor vlog. Each row contains eight full frames in temporal order. Reference-image insets and audience text in the vlogging frames are retained from the source video.}
    \label{fig:chatting-vlogging}
\end{figure}

FlowAct-R2 generates real-time 720p video. We demonstrate changing references and audience interaction in four settings. Entertainment streams combine audience conversation with singing, dancing, and other skills. Shopping streams show product presentation, item try-on, and transitions between product references. Video chats demonstrate listening and expressive turn-taking. Live vlogs combine scene changes with audience-driven branching. These examples illustrate how the same agent--generator interface supports both conversational and task-oriented behavior.
Figures~\ref{fig:performance-shopping} and~\ref{fig:chatting-vlogging} show selected frames from the demonstrations.

\subsection{Comparison with Vidu-S1.}
Vidu S1 supports real-time interactive video generation through spoken interaction~\citep{zhang2026vidus1}. In the human evaluation, FlowAct-R2 achieves GSB scores of +54.76\% for video quality and +40.48\% for real-time interaction against Vidu-S1. The accompanying comparison describes improvements in visual clarity, motion naturalness and expressiveness, action responsiveness, and the contextual appropriateness of dialogue. 

\begin{figure}[H]
    \centering
    \includegraphics[width=\linewidth,height=0.64\textheight,keepaspectratio]{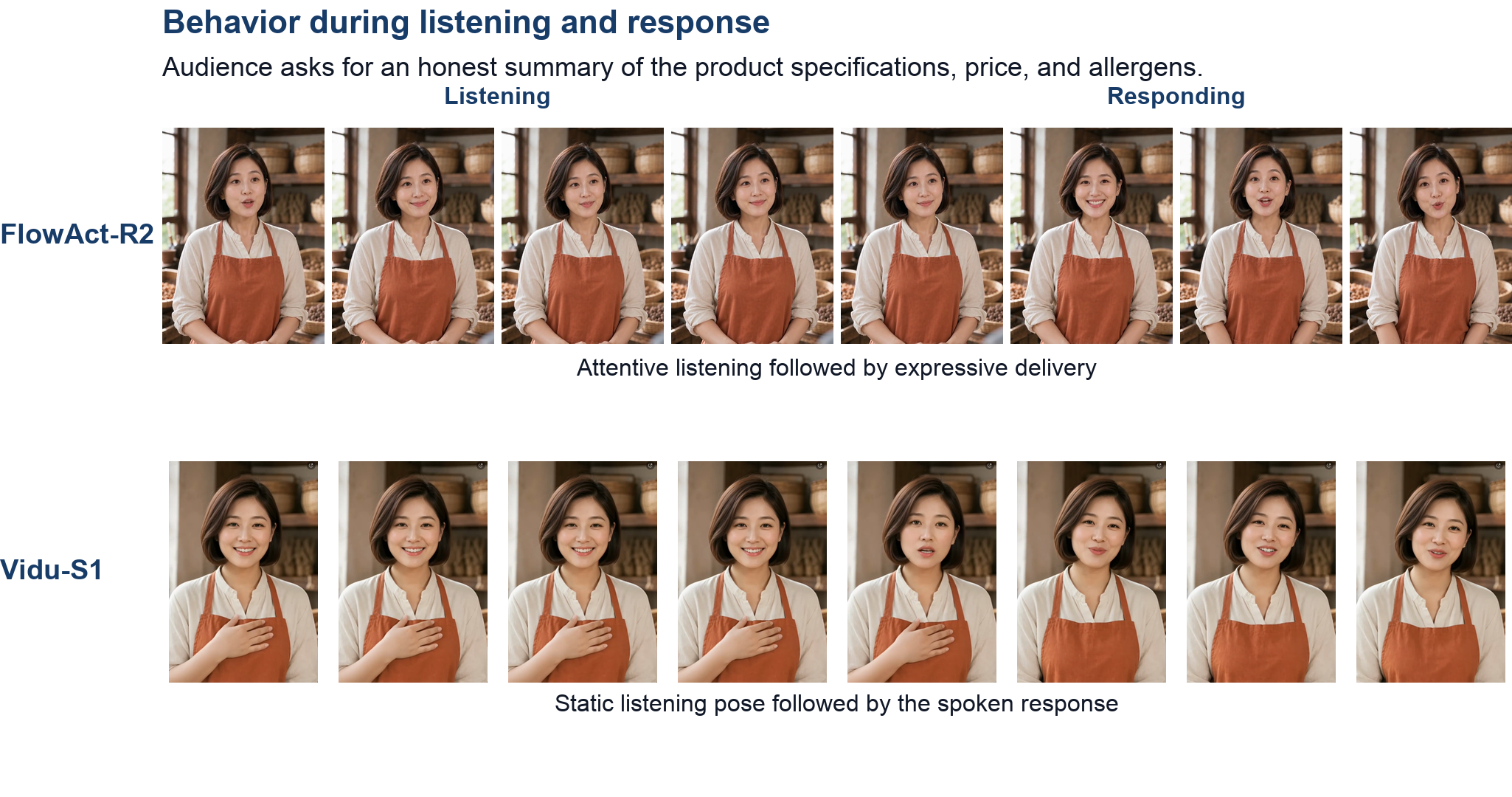}
    \caption{Behavior against Vidu-S1 during the same audience request and subsequent response. The first four columns show listening; the final four show speaking. FlowAct-R2 varies gaze, head pose, and facial response during listening, whereas Vidu-S1 maintains a nearly unchanged pose, smile, and hand gesture.}
    \label{fig:s1-action-comparison}
\end{figure}

Figure~\ref{fig:s1-action-comparison} compares the two systems across both listening and response phases. While the audience is speaking, FlowAct-R2 produces continuous listening-state behavior through changes in gaze, head pose, and facial expression, whereas Vidu-S1 remains close to the same frontal pose, smile, and hand gesture. During the subsequent response, FlowAct-R2 continues with varied facial delivery. The comparison highlights the richer nonverbal behavior planned by FlowAct-R2 across conversational turns.

\subsection{Qualitative comparison with Vidu-S2.}
\begin{figure}[H]
    \centering
    \includegraphics[width=\linewidth]{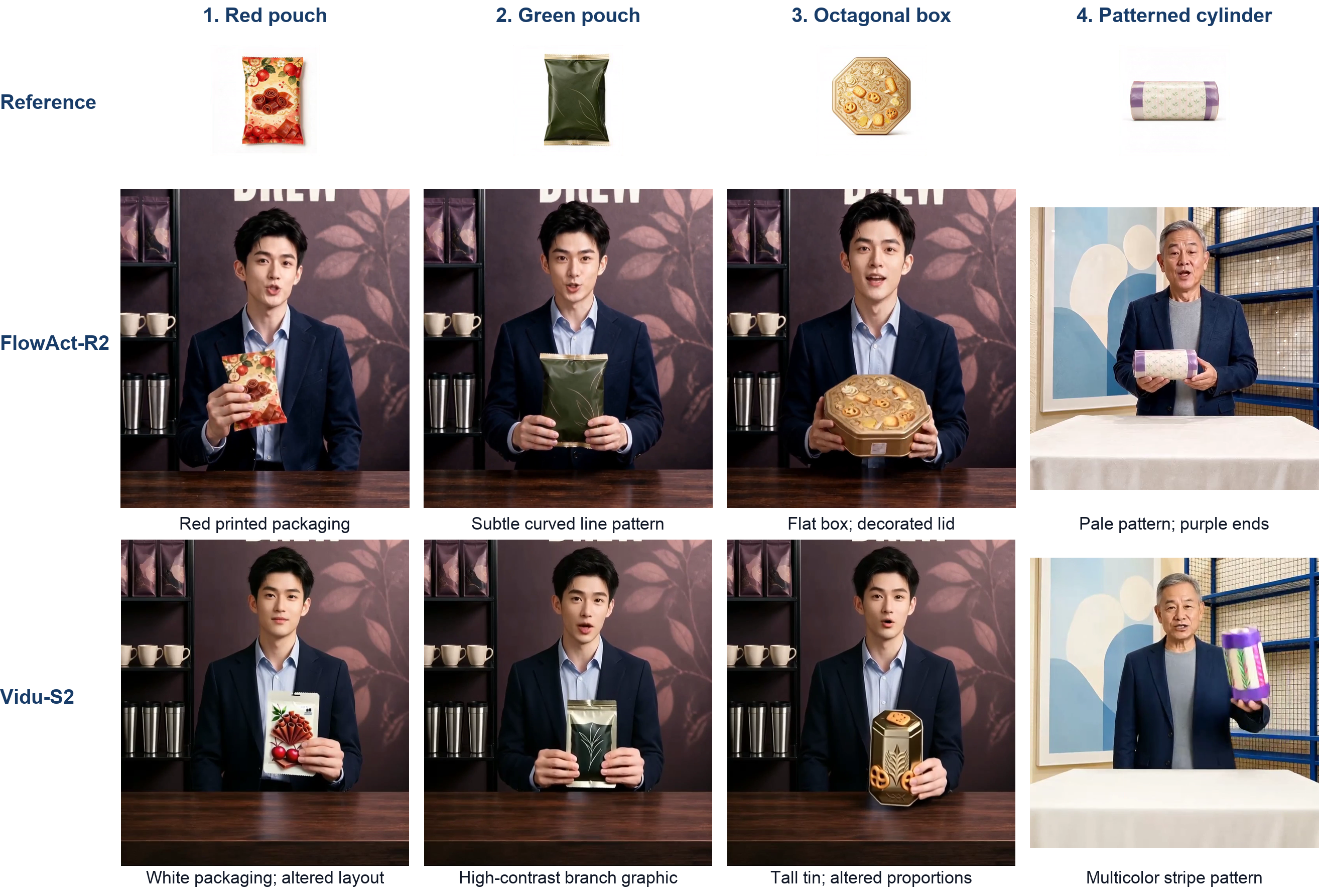}
    \caption{Product-reference fidelity against Vidu-S2. Columns 1--3 follow successive product references in one paired shopping demonstration; column 4 shows a separate cylinder case. Reference images are copied from the source-video insets. Presenter crops expose differences in package graphics and shape.}
    \label{fig:s2-reference-comparison}
\end{figure}

Vidu-S2 supports streaming reference conditioning, allowing product images to be introduced or switched during an ongoing interactive video session~\citep{zhang2026vidus2}. Figure~\ref{fig:s2-reference-comparison} compares outputs for the same product references. Across three successive products, FlowAct-R2 retains the red packaging layout, the green pouch's subtle line pattern, and the shallow octagonal box. Vidu-S2 does not complete these reference switches cleanly: attributes from adjacent products remain and blend into the newly generated object, producing predominantly white packaging, a prominent branch graphic, and a tall polygonal tin. In the separate cylinder example, it replaces the pale patterned wrap with multicolored stripes. These cases reveal both object fusion during switching and reduced fidelity in packaging details and proportions. We also demonstrate text-triggered dancing in both systems and reference-video-driven dancing in FlowAct-R2.

\section{Conclusion}
\label{sec:conclusion}

FlowAct-R2 advances live digital humans along three dimensions. First, streaming multimodal reference generation accepts rolling prompts and dynamically updated image, audio, and video references for continuous real-time video synthesis. Second, the proactive interaction agent separates offline preparation of personas, agendas, and skills from online behavior scheduling, enabling autonomous activity while responding to user input and interruptions. Third, the unified system supports both expressive and task-oriented behaviors across video chatting, live shopping, entertainment streaming, and live vlogging. Together, these capabilities move digital humans beyond reactive talking avatars toward persistent visual agents that can plan and act through continuous video.

\bibliographystyle{plainnat}
\bibliography{report/reference-review/references-candidates}
\end{document}